\pdfoutput=1

\documentclass[11pt]{article}

\usepackage[final]{acl}

\usepackage{times}
\usepackage{latexsym}

\usepackage[T1]{fontenc}

\usepackage[utf8]{inputenc}

\usepackage{microtype}

\usepackage{inconsolata}

\usepackage{graphicx} 
\usepackage{amsmath} 
\usepackage{amsfonts} 
\usepackage{amssymb} 
\usepackage{algorithm}
\usepackage{algorithmicx}
\usepackage{algpseudocode} 
\usepackage{graphicx}
\usepackage{tcolorbox}
\usepackage{enumitem}
\usepackage{subfigure}   
\usepackage{float}  
\usepackage{bm}
\usepackage{multirow}
\usepackage{xspace}
\usepackage{enumitem}
\usepackage{appendix}
\usepackage{hyperref}
\usepackage{xcolor}
\usepackage{graphicx}
\usepackage{subcaption}
\usepackage{booktabs} 
\usepackage{multirow}
\usepackage{enumitem}
\usepackage{balance}
\usepackage{makecell}
\usepackage{threeparttable}
\usepackage{amsmath,amsfonts,mathtools}
\usepackage{mathrsfs}
\usepackage{float}
\usepackage{graphicx}
\usepackage{makecell}
\usepackage{booktabs}
\usepackage{pifont}

\usepackage{multirow}
\usepackage{enumitem}
\usepackage{balance}
\usepackage{threeparttable}
\usepackage{amsmath,amsfonts,mathtools} 
\usepackage{colortbl}
\usepackage{mathrsfs}
\usepackage{float}
\usepackage{graphicx}
\usepackage{subfigure}
\usepackage{hyperref}
\usepackage{tabularx, booktabs}
\usepackage{tikz}
\usepackage{arydshln}
\usepackage{CJKutf8}
\usepackage{amsmath,lipsum}
\usepackage{cuted}
\usepackage{multirow}
\usepackage{array}
\usepackage{graphicx}
\usepackage{booktabs}

\usepackage{newfloat}
\usepackage{listings}

\newcommand{\M}{{\textsc{GRait}}\xspace}

\title{\M: Gradient-Driven Refusal-Aware Instruction Tuning \\for Effective Hallucination Mitigation}

\author{
 \textbf{Runchuan Zhu\textsuperscript{2}\thanks{Equal contribution.}},
 \textbf{Zinco Jiang\textsuperscript{2}\footnotemark[1]},
 \textbf{Jiang Wu\textsuperscript{1}\footnotemark[1]\thanks{Project lead.}},
 \textbf{Zhipeng Ma\textsuperscript{3}},
 \\
 \textbf{Jiahe Song\textsuperscript{2}},
 \textbf{Fengshuo Bai\textsuperscript{4}},
 \textbf{Dahua Lin\textsuperscript{1,5}},
 \textbf{Lijun Wu\textsuperscript{1}},
 \textbf{Conghui He\textsuperscript{1}\thanks{Corresponding author.}}
\\
 \textsuperscript{1}Shanghai Artificial Intelligence Laboratory, 
 \textsuperscript{2}Peking University, \\
 \textsuperscript{3}Southwest Jiaotong University,
 \textsuperscript{4}Shanghai Jiaotong University, \\
 \textsuperscript{5}Chinese University of Hong Kong
\\
 \small{
   \textbf{Correspondence:} \href{heconghui@pjlab.org.cn}{heconghui@pjlab.org.cn}
 }
}

\begin{document}
\maketitle
\begin{abstract}
Refusal-Aware Instruction Tuning (RAIT) aims to enhance Large Language Models (LLMs) by improving their ability to refuse responses to questions beyond their knowledge, thereby reducing hallucinations and improving reliability.
Effective RAIT must address two key challenges: firstly, effectively reject unknown questions to minimize hallucinations; secondly, avoid over-refusal to ensure questions that can be correctly answered are not rejected, thereby maintain the helpfulness of LLM outputs.
In this paper, we address the two challenges by deriving insightful observations from the gradient-based perspective, and proposing the \textbf{\underline{G}}radient-driven \textbf{\underline{R}}efusal-\textbf{\underline{A}}ware \textbf{\underline{I}}nstruction \textbf{\underline{T}}uning Framework (\textbf{\M}):
\M (1) employs gradient-driven sample selection to effectively minimize hallucinations and (2) introduces an adaptive weighting mechanism during fine-tuning to reduce the over-refusal.
Experiments on open-ended and multiple-choice question answering tasks demonstrate that \M significantly outperforms existing RAIT methods in the overall performance.
The source code and data will be available at \url{https://github.com/opendatalab/GRAIT}.

\end{abstract}

\section{Introduction}
\label{sec:Introduction}
Large Language Models (LLMs), including notable turbos like GPTs~\cite{ChatGPT,OpenAI2023GPT4TR} and LLaMA~\cite{touvron2023llama,dubey2024llama}, have achieved remarkable advances, demonstrating exceptional capabilities across a diverse range of downstream tasks~\cite{kaplan2020scaling,vu2024gptvoicetasker,achiam2023gpt,bai2024rat,jiang2024hykge,jiang2024tc}. Despite this success, critical challenges persist, particularly in the generation of \textit{\textbf{hallucinations}}—the models generate incorrect or fabricated information when confronted with unfamiliar or ambiguous queries~\cite{ji2023survey,dont_hallucinate_abstain,kang2024unfamiliar}, which ultimately limits the reliability and usefulness of LLMs.

\begin{figure}[t]
    \centering
    \includegraphics[width=1.0\linewidth]{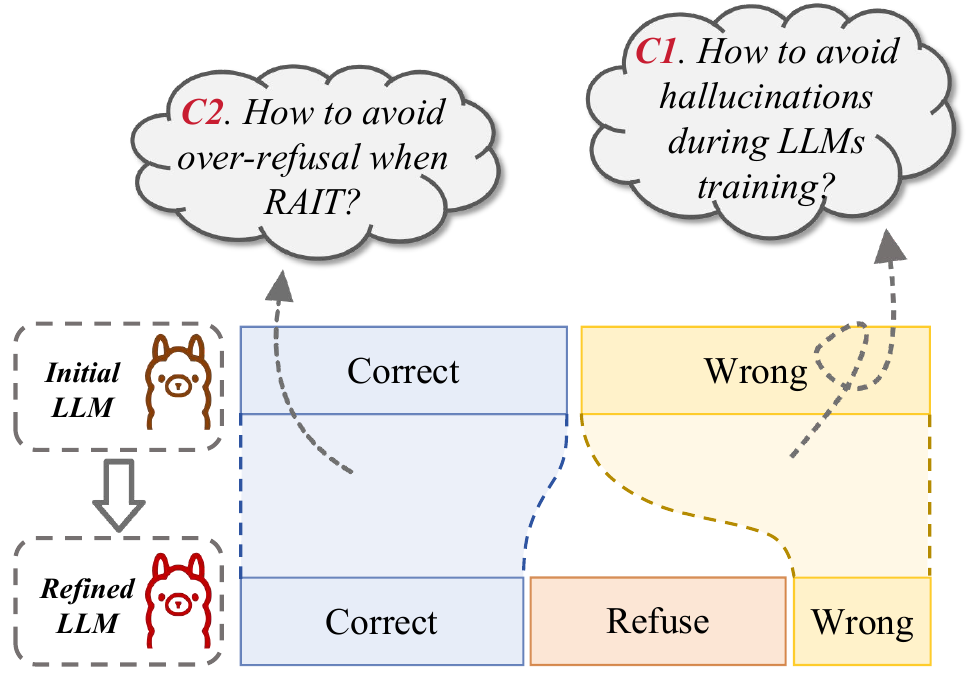}
    \caption{Descriptions of \textbf{\textit{C1} \& \textit{C2}}. After RAIT, the initial LLM model will largely reject unknown questions to avoid errors. However, the overly conservative nature of RAIT also led to a decrease in accuracy.}
    \label{fig:introduction}
\end{figure}

\begin{figure}[t]
    \centering
    \includegraphics[width=1.0\linewidth]{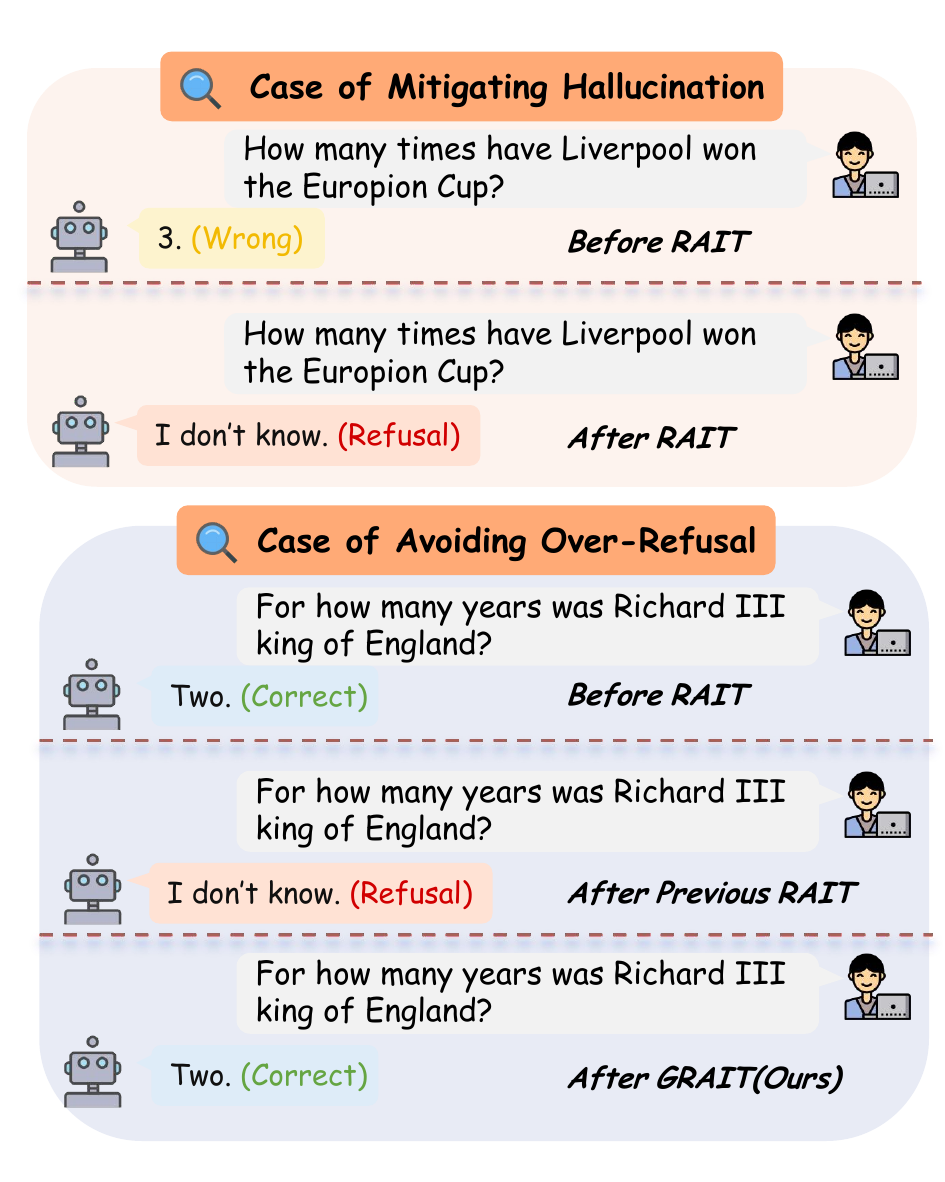}
    \caption{Case of mitigating hallucination and avoiding over-refusal.}
    \label{fig:case}
\end{figure}

Ideally, the responsible LLM should decline to answer questions beyond its knowledge to minimize hallucinations~\cite{know_your_limits,survey_honesty}.
Recent studies~\cite{alignment_for_honesty, R_Tuning, rejection_improves, cheng2024can, bai2024efficient, zhu2024utilizeflowsteppingriver} have developed Refusal-Aware Instruction Tuning (RAIT), which constructs the refusal-aware dataset and employs Supervised Fine-Tuning (SFT)~\cite{dongabilities,Ouyang_Wu_Jiang_Almeida_Wainwright_Mishkin_Zhang_Agarwal_Slama_Ray_et,luo2024kuaiji} to teach models to appropriately decline responses.
Typically, refusal-aware datasets~\cite{R_Tuning,zhangdefending} categorize training samples into \texttt{ik} (correct) and \texttt{idk} (incorrect) groups based on response correctness. Samples with incorrect responses (\texttt{idk}) are treated as unknown knowledge and the answers are replaced with refusal responses like ``I don't know'', while the correct (\texttt{ik}) remain unchanged. Despite RAIT’s success in reducing hallucinations~\cite{R_Tuning,alignment_for_honesty,zhang2025amulet,wan2024mitigating}, studies like~\cite{varshney2023art} and~\cite{cheng2024can} highlight that models can become overly cautious, leading to over-refusals. 
Therefore, as shown in Figure \ref{fig:introduction} and Figure \ref{fig:case}, the RAIT should address the following two challenges \textit{\textbf{simultaneously}}:
\textbf{\textit{C1}. How to effectively reduce the hallucinations by refusing the unknown questions?}
\textbf{\textit{C2}. How to avoid over-refusal to ensure questions that can be correctly answered are not rejected?}

To address these challenges, we introduce the \textbf{\underline{G}}radient-based \textbf{\underline{R}}efusal-\textbf{\underline{A}}ware \textbf{\underline{I}}nstruction \textbf{\underline{T}}uning Framework~(\textbf{\M}), which has several advantages over previous RAIT methods:
(1) Unlike prior methods that rely solely on the outputs of LLMs, \M utilizes gradients to achieve a more accurate representation of LLMs' internal knowledge states, ensuring better alignment between the constructed training samples and the LLMs' knowledge.
(2) Our gradient-based sample selection process is more efficient, achieving comparable training results with fewer samples \cite{Xia_Malladi_Gururangan_Arora_Chen, xu2024parenting}.
(3) By incorporating gradient information, we account for influences among training samples \cite{ren2024learningdynamics}, effectively minimizing sample conflicts \cite{zhu2024utilizeflowsteppingriver} and better addressing the aforementioned two challenges.

In practical implementation, we first derive two theoretical observations ($\mathbf{O_1}$ \& $\mathbf{O_2}$) by progressively addressing reduced inaccuracies (\textit{\textbf{C1}}) and mitigating over-refusal (\textit{\textbf{C2}}), which form the basis for designing \M. 
Then in the framework designing: \ding{182} For \textbf{\textit{C1}}, we leverage the Refusal Influence formula within the \texttt{idk} set to select a small subset, enabling the LLMs to learn the refusal paradigm while filtering out inefficient samples.
\ding{183} For \textbf{\textit{C2}}, we implement an adaptive weighting method derived from the Stable Influence formula between the correct and incorrect sample sets, assigning varying sample weights during the RAIT phase to alleviate the issue of over-refusal. In summary, our contributions are as follows:

\begin{itemize}[leftmargin=*,noitemsep,topsep=2pt]
\item To the best of our knowledge, we are the first to conduct a theoretical analysis of the causes underlying the over-refusal phenomenon in LLMs.
\item The \M framework establishes a comprehensive workflow encompassing data construction and fine-tuning, utilizing two gradient-driven observations to enhance the model's refusal capability while effectively mitigating over-refusal.
\item Through extensive experimental evaluation on both open-ended question answering and multiple-choice tasks, we demonstrate that \M surpasses existing baselines by significantly reducing hallucination rates and enhancing overall performance.
\end{itemize}

The paper is organized as follows. In Introduction (c.f. Section~\ref{sec:Introduction}), we present our research question and its corresponding two challenges (\textit{\textbf{C1}} \& \textit{\textbf{C2}}). The Related Work (c.f. Section~\ref{sec:Related Work}) provides a literature analysis concerning each challenge. In the Preliminary section (c.f. Section~\ref{sec:Preliminary}), we define the symbols and tasks relevant to \M. Following this, the Theoretical Analysis section (Section~\ref{sec:Theoretical Analysis}) derives two observations, $\mathbf{O_1}$ \& $\mathbf{O_2}$, in a progressive manner corresponding to \textit{\textbf{C1}} \& \textit{\textbf{C2}}. The Method section (Section~\ref{sec:Methodology}) aligns $\mathbf{O_1}$ with data construction and $\mathbf{O_2}$ with the RAIT phase. Section \ref{sec:Experiment} introduces the experiment settings and the experimental results. Finally, Section \ref{sec:Conclusion} concludes this paper and discusses future research directions.

\section{Related Work}
\label{sec:Related Work}
\subsection{Refusal-Aware Instruction Tuning}
RAIT is a supervised technique designed to enhance the ability of LLMs to handle unanswerable or uncertain questions by training them to respond directly with ``I don't know''~\cite{R_Tuning,alignment_for_honesty,zhu2024utilizeflowsteppingriver} through supervised fine-tuning (SFT). 
In \citet{wan2024mitigating}, a knowledge-based verification mechanism is proposed to ensure that the model's knowledge remains consistent with external trusted sources to prevent the spread of misinformation.
Moreover, CoKE~\cite{cokeee} probes LLMs' knowledge boundaries via internal confidence given a set of questions and then informs the LLM's decision on whether to respond with ``I don't know'' based on the knowledge boundaries. Additionally, \citep{zhu2024utilizeflowsteppingriver} and \citep{ula} refine data filtering and modification by leveraging both response certainty and correctness.
Recent works have incorporated Low-Rank Adaptation (LoRA)~\cite{hulora} and AdaLoRA~\cite{Wolfe_2024} into RAIT to achieve further improvements. However, those RAIT methods tend to make LLMs more conservative, leading to incorrect and even \textbf{over-refusals}~\cite{cheng2024can}.

\subsection{Gradient Effect on LLMs' Learning}
Gradient-based methods are central to recent advances in data selection. For instance, \citep{zhao2021dataset,Xia_Malladi_Gururangan_Arora_Chen,yang2024smalltolarges2lscalabledata} propose \textit{Dataset Condensation}, where synthetic data is created by matching the gradients or the learning trajectory of a deep model trained on a small synthetic set to those of a model trained on the full dataset. Similarly, \citep{killamsetty2021gradmatch,killamsetty2021subset} extend this concept with their \textit{Grad-Match} framework, which selects subsets of data that closely align with the gradient information from the full dataset, allowing for efficient training with minimal performance degradation.
More recently, \citep{ren2024learningdynamics,zhao2020dataset,qiao2024PGP,Bai_Zhang_Tao_Wu_Wang_Xu_2023} emphasize that gradients can dynamically influence the learning process of LLMs during fine-tuning. 
\citep{Liu_Jiang_Bai_Chen_Wang_2020,Xiao2024CertifiedRobustness} study the influence of gradient signal-to-noise ratio's result on the test set.
While gradient-based selection techniques are widely studied, applying these methods to RAIT to mitigate issues such as hallucinations and over-refusal remains largely unexplored, which presents an opportunity for further research in RAIT.

\section{Preliminary}
\label{sec:Preliminary}
\paragraph{\textit{(Definition 1. RAIT Dataset)}} 
The RAIT process can be described as follows: the initial LLM is prompted to answer all questions in the training set $D_{\text{src}}$. Based on the correctness of the responses, the samples are categorized into two groups. 
\textit{\textbf{\ding{182}}} Samples with correct responses are considered known knowledge of the LLM. These answers will remain unchanged and are referred to as \texttt{ik} samples, denoted as $D_{\text{ik}} = \{(x_{\text{ik}}, y_{\text{ik}})\}$ (where `ik' stands for `I know', $x_{\text{ik}}$ is the known question and $y_{\text{ik}}$ is the ground-truth label). \textit{\textbf{\ding{183}}} Conversely, samples with incorrect responses are treated as unknown knowledge. Their original answers are replaced with refusal responses such as ``I don't know'' forming $D_{\text{idk}} = \{(x_{\text{idk}}, y_{\text{idk}})\}$ (where `idk' stands for `I don't know', $x_{\text{ik}}$ is the unknown question and $y_{\text{ik}}$ is modified refusal response such as ``I don't know''). 
The constructed RAIT dataset, $D_{\text{rait}} = D_{\text{ik}} \cup D_{\text{idk}}$, is used to fine-tune the initial LLM, parameterized by $\theta$, to improve its ability to refuse to answer questions beyond its knowledge.

\paragraph{\textit{(Definition 2. Influence Formulation)}}
To estimate the influence of a training datapoint on a validation sample, we use the first-order Taylor expansion of the loss function \cite{Pruthi_Liu_Kale_Sundararajan_2020}\footnote{The reasons for using the influence formula are outlined in the appendix \ref{app:Reasons for Choosing Influence Formula}}. Specifically, for a model $\theta_t$ at step $t$, the loss on unobservant validation sample $x^{u}$ can be approximated as:
$
\mathcal{L}(x^{u},y^{u}; \theta_{t+1}) \approx \mathcal{L}(x^{u},y^{u}; \theta_t) + \langle \nabla \mathcal{L}(x^{u},y^{u}; \theta_t), \theta_{t+1} - \theta_t \rangle.
$
If the model is trained using Stochastic Gradient Descent (SGD) with batch size 1 and learning rate $\eta_t$, for the observant training sample $x^o$, the SGD update is written as:
$
\theta_{t+1} - \theta_t = -\eta_t \nabla \mathcal{L}(x^{o},y^{o}; \theta_t).
$
At this point, we can define the influence formula of $(x^{o},y^{o})$:
\begin{equation}
\small
\label{eq:influence_equation}
\begin{aligned}
\mathcal{I}(x^{o},y^{o},x^{u},y^{u}; \theta_{t}) \stackrel{\triangle}{=} & ~ \eta_t \langle \nabla \mathcal{L}(x^{o},y^{o}; \theta_t), \\ \quad \quad
& \nabla \mathcal{L}(x^{u},y^{u}; \theta_t) \rangle.
\end{aligned}
\end{equation}

\paragraph{\textit{(Task Definition)}}
The objective of this task is to leverage $D_{\text{rait}}$ to fine-tune a model and minimize the loss on two distinct types of test samples. Specifically, for samples that were previously incorrect, we aim for the model to output answers like ``I don't know'', while for correct samples, the predicted label should be as close as possible to the ground-truth label $y_{\text{ik}}$
. The task can be formalized as minimizing the following loss:
\begin{equation}
\small
\begin{aligned}
\label{eq:task}
\min \bigl \{ \mathbb{E}_{x^{u}_{{\text{idk}}} \sim D_{{\text{idk}}}} &\left[ \Delta \mathcal{L}(x^{u}_{{\text{idk}}}, y^{u}_{{\text{idk}}}; \theta) \right ]\\ +& \mathbb{E}_{x^{u}_{{\text{ik}}} \sim D_{{\text{ik}}}} \left[ \Delta \mathcal{L}(x^{u}_{{\text{ik}}}, y^{u}_{{\text{ik}}}; \theta) \right ] \bigl\},
\end{aligned}
\end{equation}
In addition to minimizing it, a key objective of this task is to select the most suitable subset $\widetilde{D}_{\text{rait}} \subseteq D_{\text{rait}}$ for fine-tuning (c.f. Section~\ref{sec:Theoretical Analysis} for proof). By selecting optimal data from the RAIT dataset, we aim to improve the model's ability to refuse answers to unknown questions while minimizing over-refusal.

\section{Theoretical Analysis}
\label{sec:Theoretical Analysis}
This part is organized as $\mathbf{O}_1 \to \mathbf{O}_2$. Before obtaining formal observation results, we first propose two assumptions:

\paragraph{\textit{Assumptions 1. Distribution Assumption}}
\textit{We assume that the distributions of \texttt{ik} or \texttt{idk} from train and test sets are identically distributed, formally expressed as:}
$
\Pi_{D_{\text{idk}}^o} \sim \Pi_{D_{\text{idk}}^{u}},  \Pi_{D_{\text{ik}}^o} \sim \Pi_{D_{\text{ik}}^{u}}
$.

\paragraph{\textit{Assumptions 2. Orthogonality of Means}}
\textit{We further assume that the means of the gradient distributions for \texttt{idk} and \texttt{ik} are orthogonal as verified in Appendix~\ref{subsec:orthogonal_experiment}, and we have:}
\begin{equation}
\small
\label{eq:loss_decomposition}
\begin{aligned}
\Bigl \langle \mathbb{E}_{x_{*} } \left[ \nabla \mathcal{L}(x_{*},y_{\text{idk}}; \theta) \right],
\mathbb{E}_{x_{*} } \left[ \nabla \mathcal{L}(x_{*},y_{\text{ik}}; \theta) \right]
\Bigr \rangle \approx 0,
\end{aligned}
\end{equation}
where the \( * \) denotes the symbol of either \texttt{idk} or \texttt{ik}.

\subsection{Reducing Incorrectness ($\mathbf{O}_1$)}
\label{sec:Reducing Incorrectness}
We begin by focusing on minimizing the loss to improve the rejection rate, specifically aiming to minimize the first term of \eqref{eq:task} $\mathbb{E}_{x^{u}_{\text{idk}} \sim D_{\text{idk}}} \left[ \Delta \mathcal{L}(x^{u}_{\text{idk}}, y^{u}_{\text{idk}}; \theta) \right]$. Then, combining equation~\eqref{eq:influence_equation}, we can express this as:
\begin{equation}
\small
\label{eq:loss_decomposition_short}
\begin{aligned}
&\mathbb{E}_{x^{u}_{\text{idk}} \sim D_{\text{idk}}} \bigl[ \Delta \mathcal{L}(x^{u}_{\text{idk}}, y^{u}_{\text{idk}}; \theta) \bigl]  
\approx 
- \mathbb{E} _{(x^{u} _{\text{idk}}, x^o _{\text{idk}}) \sim D _{\text{idk}}} \\  &\left[ \mathcal{I}(x^{o} _{\text{idk}}, y^{o} _{\text{idk}}, x^{u} _{\text{idk}}, y^{u} _{\text{idk}}; \theta) \right]
\end{aligned} \noindent
\end{equation}
and the full proof is detailed in Appendix~\ref{app:More Proof on O1}.

Thus, samples with gradients similar to the average gradient direction of $D_{\text{idk}}$ are the most effective in reducing the model's hallucination rate.

\subsection{Alleviating Over-Refusal ($\mathbf{O}_2$)}
\label{sec:Alleviating Over-Refusal}
However, we observed that if we optimize the model merely depends on RAIT,
it leads to the issue of \textbf{over-refusal} (i.e., \texttt{ik} samples also tend to output ``I don't know''). Therefore, we delved deeper into the whole target in \eqref{eq:task} and derived the following( the full proof is detailed in Appendix \ref{app:More Proof on O2}):
\begin{equation}
\small
\label{eq:O2}
\begin{aligned}
&\mathbb{E}_{x^{u}_{\text{idk}} \sim D_{\text{idk}}} \left[ \Delta \mathcal{L}(x^{u}_{\text{idk}}, y^{u}_{\text{idk}}; \theta) \right ] 
+ \mathbb{E}_{x^{u}_{\text{ik}} \sim D_{\text{ik}}} \left[ \Delta \mathcal{L}(x^{u}_{\text{ik}}, y^{u}_{\text{ik}}; \theta) \right ] \\ 
\approx & - \left \{ \mathbb{E} _{x^{u} _{idk}, x^o _{idk} \sim D _{idk}} \left[ \mathcal{I}(x^{o} _{\text{idk}}, y^{o} _{\text{idk}}, x^{u} _{\text{idk}}, y^{u} _{\text{idk}}; \theta) \right] \right.  -\\
\quad &  \left . \mathbb{E} _{x^{u} _{\text{ik}} \sim D _\text{{ik}}, x^{o} _{\text{idk}} \sim D _{\text{idk}}} \left[ \mathcal{I}(x^{o} _{\text{idk}}, y^{o} _{\text{idk}}, x^{u} _{\text{ik}}, y^{u} _{\text{idk}}; \theta) \right] \right \}
\end{aligned} \noindent
\end{equation}

The first expectation term in equation \eqref{eq:O2} captures the reduction in the model's error rate, while the second term reflects the occurrence of over-refusal. Training samples where the difference between these two terms is smaller tend to exacerbate over-refusal, though they may also contribute to stronger overall model performance.

\section{Methodology}
\label{sec:Methodology}
\begin{figure*}[t]
  \centering
  \includegraphics[width=1.0\linewidth]{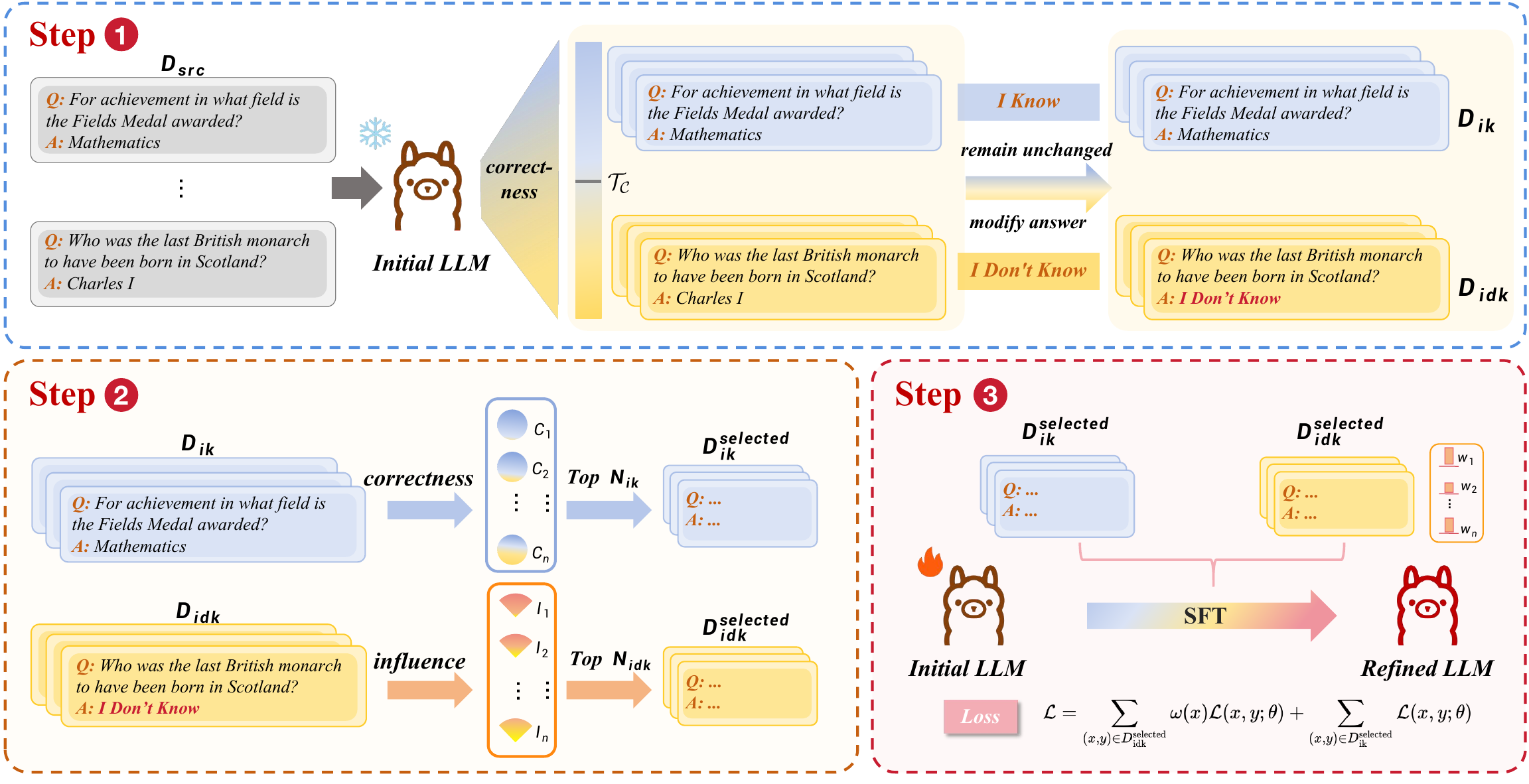}
  \caption{\textbf{Overview of our framework.} \M contains three stages: 
  \textbf{(1)} Constructing datasets $\mathbf{D_{\text{ik}}}$ and $\mathbf{D_{\text{idk}}}$ by querying the internal state of LLMs.
  \textbf{(2)} Distilling the datasets to select \texttt{idk} samples based on the first observation $\mathbf{O}_1$.
  \textbf{(3)} Performing Influence-directed Refusal-aware Instruction Tuning using the second observation $\mathbf{O}_2$.
  }
  \label{fig:method}
\end{figure*}

In this Section, we follow the two observations ($\mathbf{O}_1$\&$\mathbf{O}_2$) in Section~\ref{sec:Theoretical Analysis} and split \M into three stages: 
 
\begin{itemize}[leftmargin=*,noitemsep,topsep=2pt]
\item \textbf{Stage 1: Construct $\mathbf{D_{\text{ik}}}$ \& $\mathbf{D_{\text{idk}}}$}, which obtain $\mathbf{D_{\text{ik}}}$ \& $\mathbf{D_{\text{idk}}}$ by querying the internal state of LLMs and modifying the label of the incorrect set to `I don't know'.

\item \textbf{Stage 2: Dataset Construction based on  $\mathbf{O}_1$}, select \texttt{idk} samples from the first observation.

\item \textbf{Stage 3: Influence-directed Refusal-aware Instruction Tuning based on $\mathbf{O}_2$}, which allocates different weight when Refusal-aware Instruction Tuning from the second observation.
\end{itemize}
Figure~\ref{fig:method} presents a detailed overview of our proposed \M~framework, Algorithm~\ref{alg:RAI_data} details the overall process with subsequent subsections detailing each component.

\subsection{Stage 1: Construct $\mathbf{D_{\text{ik}}}$ \& $\mathbf{D_{\text{idk}}}$}
In the first stage, we compute the accuracy \(\mathcal{C}(x)\) for each sample \(x\). Our research focuses on two distinct tasks: Multiple-Choice Question Answering (MCQA) and Open-Ended Question Answering (OEQA). For the MCQA task, we use the token probability of the ground truth to evaluate the correctness of each sample. In the OEQA task, the model generates answers \(N\) times (with \(N = 10\)) for each question, and we calculate accuracy based on the generated responses. 

Next, we construct the dataset \(\mathbf{D_{\text{ik}}}\) using samples with \(\mathcal{C}(x)\) of the correctness threshold $\mathcal{T}_{\mathcal{C}}$ or higher. For samples with \(\mathcal{C}(x)\) below $\mathcal{T}_{\mathcal{C}}$, we modify their output to ``I don't know'' and construct these samples into a separate dataset, \(\mathbf{D_{\text{idk}}}\).

\subsection{Stage 2: Dataset Construction based on  $\mathbf{O}_1$}
In this module, as outlined in Algorithm\ref{alg:RAI_data}, we select the most efficient data to train the model. \textbf{First, for the \texttt{idk} set $\mathbf{D_{\text{idk}}}$}, we identify the most impactful \texttt{idk} samples that significantly enhance the model's ability to refuse unknown questions based on the conclusion of $\mathbf{O}_1$. Specifically, we approximate \eqref{eq:loss_decomposition_short} and define \textbf{Refusal Influence} of each training sample on the model’s loss as:

\begin{equation} 
\label{eq} 
\small 
\mathcal{I}^{\text{ref}}(x) = \Bigl \langle \nabla \mathcal{L}(x; \theta), \mathbb{E}_{x^{o} \sim D_{\text{idk}}} \left[ \nabla \mathcal{L}(x^{o}_{\text{idk}}, y^{o}_{\text{idk}}; \theta) \right] \Bigr \rangle,
\end{equation}
where \(x\) denotes the training sample, \(\mathcal{L}\) is the loss function, and \(\theta\) represents the model parameters. This influence metric allows us to assess the contribution of each \texttt{idk} sample towards minimizing the incorrectness.
Building upon the influence scores, we employ a ranking strategy to select the most influential \texttt{idk} samples for fine-tuning:
\begin{equation}
\small
\label{eq:selected}
\mathbf{D_{\text{idk}}^{\text{selected}}} = \{ x \in \mathbf{D_{\text{idk}}} \mid \texttt{Rank}\bigl( \mathcal{I}^{\text{ref}}(x) \bigl) \leq N_{\text{idk}} \}.
\end{equation}
The selected \texttt{idk} samples, which exhibit the highest refusal influence, will form a distilled dataset used for targeted training.
Furthermore, to ensure computational efficiency, we apply two techniques following~\cite{Xia_Malladi_Gururangan_Arora_Chen} to construct valuable low-dimensional gradient features: parameter efficient fine-tuning via LoRA~\cite{hu2021lora} and random projections~\cite{projection}.

\textbf{Secondly, for the dataset $\mathbf{D_{\text{ik}}}$}, we select the subset of data with the highest accuracy. Previous work~\cite{ren2024learning_or_align} has demonstrated that using this subset for fine-tuning does not negatively impact the model's overall performance. The selection of this data can be represented as:
\begin{equation}
\small
\label{eq:selected}
\mathbf{D_{\text{ik}}^{\text{selected}}} = \{ x \in \mathbf{D_{\text{ik}}} \mid \texttt{Rank}\bigl( \mathcal{C}(x) \bigl) \leq N_{\text{ik}} \}.
\end{equation}

\subsection{Stage 3: Influence-directed Refusal-aware Instruction Tuning based on $\mathbf{O}_2$}
In this stage, we introduce an Influence-directed Refusal-aware Instruction Tuning based on the conclusions from $O_{2}$, which mentions that the issue of Over-Refusal is closely related to the difference in influence between training samples. The larger this difference, the more stable the accuracy of the model remains when learning the ability to reject. Therefore, we propose the concept of \textbf{Stable Influence} and assign each sample a weight \(\omega\), which is calculated as follows:
\begin{equation}
\small
\label{eq:loss_decomposition}
\begin{aligned}
\mathcal{I}^{\text{sta}}(x) = & \Bigl \langle \nabla \mathcal{L}(x; \theta), 
  \mathbb{E}_{x^{o} \sim D_{\text{idk}}} \left[ \nabla \mathcal{L}(x^{o}_{\text{idk}}, y^{o}_{\text{idk}}; \theta) \right] \\
  & - \mathbb{E}_{x^{o} \sim D_{\text{ik}}} \left[ \nabla \mathcal{L}(x^{o}_{\text{ik}}, y^{o}_{\text{idk}}; \theta) \right] 
  \Bigr \rangle,
\end{aligned}
\end{equation}
\vspace{-0.3cm}
\begin{equation}
\small
\label{eq:loss_decomposition}
\begin{aligned} 
  \omega(x^{o}) = & \frac{e^{\mathcal{I}^{\text{sta}}(x^{o}) / \tau}}{\mathbb{E}_{x^{o} \sim D^{\text{selected}}_{\text{idk}}} \left[ e^{\mathcal{I}^{\text{sta}}(x^{o}) / \tau} \right]}, 
\end{aligned}
\end{equation}
where \(x^{o}\) represents a training sample, and \(\tau\) refers to the temperature parameter. As \(\tau\) approaches 0, the differences in weight distribution become more pronounced, while as \(\tau\) approaches 1, the changes in weights become minimal. The constraint \(\sum \omega(x^o) = 1\) ensures that the model maintains its ability to reduce error rates.

Finally, during training, we apply a weight to the loss of the \texttt{idk} samples to mitigate over-refusal. The SFT loss is calculated as follows:

\begin{equation}
\small
\label{eq:loss_decomposition}
\begin{aligned}
\mathcal{L}_{SFT} = & \sum\limits_{(x^o, y^o) \in D^{\text{selected}}_{\text{idk}}} \omega(x^o) \mathcal{L}(x, y^o; \theta)\\
&+ \sum\limits_{(x^o, y^o) \in D^{\text{selected}}_{\text{ik}}} \mathcal{L}(x^o, y^o; \theta).
\end{aligned}
\end{equation}

\begin{algorithm}[t]
\caption{\M Process}
\label{alg:RAI_data}
\textbf{Input:}{~$D_{\text{src}} = \{x_{0}, x_{1}, ..., x_{N}\}$}, $\mathcal{T}_{\mathcal{C}}$, $\tau$ $N_{\text{ik}}$, $N_{\text{idk}}$ \\
\textbf{Output:}{~$D_{rait}$}
\begin{algorithmic}[1]
\State $D_{\text{ik}} = \{x \, | \, x \in D_{\text{src}},  \mathcal{C}(x) \geq \mathcal{T}_{\mathcal{C}}\}$
\State $D_{\text{idk}} = \{x \, | \, x \in D_{\text{src}},  \mathcal{C}(x) < \mathcal{T}_{\mathcal{C}}\}$
\State $\overline{g}_{\text{idk}}(D_{\text{idk}}) = \frac{1}{|D_{\text{idk}}|} \sum_{x \in D_{\text{idk}}} x.g_{\text{idk}}$
\State $\overline{g}_{\text{idk}}(D_{\text{ik}}) = \frac{1}{|D_{\text{ik}}|} \sum_{x \in D_{\text{ik}}} x.g_{\text{idk}}$
\For{$x_i$ in $D_{\text{idk}}$}
    \State $x_i.\mathcal{I}^{\text{ref}} = x_i.g_{\text{idk}} \cdot \overline{g}_{\text{idk}}(D_{\text{idk}})$
    \State $x_i.\mathcal{I}^{\text{sta}} = x_i.g_{\text{idk}} \cdot [\overline{g}_{\text{idk}}(D_{\text{idk}}) - \overline{g}_{\text{idk}}(D_{\text{ik}})]$
\EndFor
\State $D_{\text{ik}} = \text{sort}(D_{\text{ik}}, \text{key}=\mathcal{C}, \text{order=descend})$
\State $D_{\text{idk}} = \text{sort}(D_{\text{idk}}, \text{key}=\mathcal{I}^{\text{ref}}, \text{order=descend})$
\State $D_{\text{ik}}^{\text{selected}} = \text{TopK}(D_{\text{ik}}, N_{\text{ik}})$
\State $D_{\text{idk}}^{\text{selected}} = \text{TopK}(D_{\text{idk}}, N_{\text{idk}})$

\State $\text{Initialize: } Z = 0$ 
\For{$x_i$ in $D_{\text{idk}}^{\text{selected}}$}
    \State $Z = Z + e^{x_i.\mathcal{I}^{\text{sta}} / \tau}$
\EndFor

\For{$x_i$ in $D_{\text{idk}}^{\text{selected}}$}
    \State $x_i.\omega_i = \frac{e^{x_i.\mathcal{I}^{\text{sta}} / \tau}}{Z/|D_{\text{idk}}^{\text{selected}}|}$ 
\EndFor

\State $D_{\text{rait}} = D_{\text{ik}} \cup D_{\text{idk}}$
\State \textbf{return} $D_{\text{rait}}$
\end{algorithmic}
\end{algorithm}

\section{Experiment}
\label{sec:Experiment}
In this section, we provide detailed information on experimental setup, and further analysis to validate the performance and rationality of \M.

\subsection{Experiment Setup}
\paragraph{Datasets.}
In this study, we assess the efficacy of \M in handling two distinct types of Question and Answering tasks: the knowledge-based Multiple Choice Question Answering (MCQA) and Open-ended Question Answering (OEQA). For the MCQA task, the test split of MMLU~\cite{MMLU} is adopted as the training dataset, while the validation split of the same serves as the In-Domain (ID) test set, and the ARC-c~\cite{ARC_C} test split is utilized as the Out-Of-Domain (OOD) test set. In the context of the OEQA task, we use the training split of TriviaQA~\cite{triviaqa} for training purposes, the development split of TriviaQA as the ID test set, and the validation split of NQ~\cite{nq} as the OOD test set. Additional information is provided in Table~\ref{table:dataset_details}.

\begingroup
\fontsize{9}{11}\selectfont
\setlength{\tabcolsep}{1mm}
\begin{table}[h]
\centering
\caption{Datasets Details.}
\resizebox{1.0\linewidth}{!}{
\begin{tabular}{lcc}
\toprule
\textbf{} & \textbf{MCQA} & \textbf{OEQA} \\ 
\midrule
\textbf{Train}     & MMLU test (14,079)     & TriviaQA train (87,622)    \\ 
\textbf{ID Eval}   & MMLU val (1,540)       & TriviaQA dev (11,313)      \\ 
\textbf{OOD Eval}  & ARC-c dev (1,172)      & NQ dev (3,610)             \\ 
\bottomrule
\end{tabular}}
\label{table:dataset_details}
\end{table}
\endgroup

\paragraph{Baselines.}
To evaluate the performance of \M, we conducted comparisons with several existing approaches:
\textbf{Init-Basic}: Employs the initial LLM setup, utilizing standard question-answering prompts to guide the model in generating answers.
\textbf{Init-Refuse}: Builds on Init-Basic by incorporating instructions such as ``\textit{If you do not know the answer, please respond with `I don't know.'}'' to promote safer responses~\cite{bianchi2024safetytunedllamaslessonsimproving,zhangdefending}.
\textbf{Van-Tuning}: Randomly selects \(N_{\text{ik}} + N_{\text{idk}}\) samples from \(D_{\text{src}}\) for straightforward instruct-tuning, without any sample modification.
\textbf{R-Tuning}: Follows the settings from \cite{R_Tuning}, where samples in the RAIT dataset are modified based on the correctness of the model's replies.
\textbf{CRaFT}: This method is implemented according to~\cite{zhu2024utilizeflowsteppingriver}, addressing both static and dynamic conflicts within the RAIT dataset to provide a thorough evaluation of potential issues.

\begin{figure}[t]
    \centering
    \includegraphics[width=0.7\linewidth]{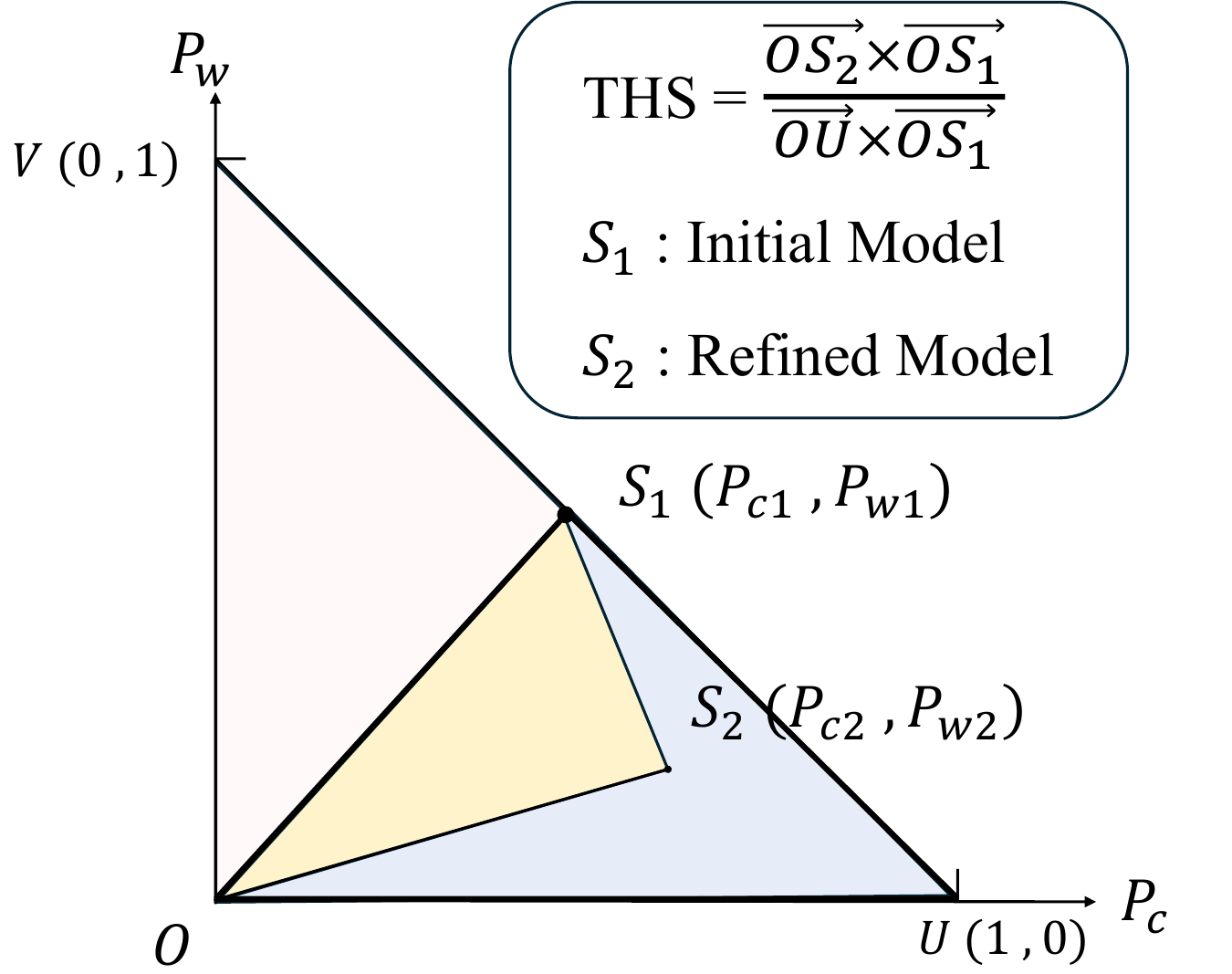}
    \caption{Illustration of Truthful Helpfulness Score.}
    \label{fig:metric}
\end{figure}

\paragraph{Evaluation Metrics.}

We utilize the Truthful Helpfulness Score (THS) as detailed by~\cite{zhu2024utilizeflowsteppingriver} to assess the performance of LLMs after RAIT. Accuracy (\(P_c\)), error rate (\(P_w\)), THS, etc. are key metrics for evaluating the performance of models after RAIT. Among these, \(P_c\) and \(P_w\) form a competing pair, where optimizing for \(P_c\) often leads to a decline in \(P_w\). Focusing on only one of these metrics is insufficient to evaluate the model’s overall capability.
Thus, a singular and comprehensive metric is required to simplify the assessment process and eliminate the complexity of balancing multiple trade-off metrics.

For each test sample, we classify the response as correct, incorrect, or refused. From these categories, we calculate the accuracy (\(P_c\)), error rate (\(P_w\)), and refusal rate (\(P_r\)). We then set up a Cartesian coordinate system with \(P_c\) and \(P_w\) on the axes. The point \(S_1\) represents the coordinates of the baseline LLM, and \(S_2\) corresponds to the refined model.
If \(S_2\) is positioned below the line from the origin \(O\) to \(S_1\) (denoted as \(OS_1\)), then a larger area of the triangle \(\triangle OS_1S_2\) signifies an improvement in the model. If, however, \(S_2\) is above \(OS_1\), it indicates a reduction in performance. As shown in Figure \ref{fig:metric}, THS is defined as the ratio of the cross product of vectors \(\overrightarrow{OS_1}\) and \(\overrightarrow{OS_2}\) to the maximum possible value of this cross product:

\begin{equation}
\small
\label{eq:loss_decomposition}
\begin{aligned}
\text{THS} = (\overrightarrow{OS_2} \times \overrightarrow{OS_1}) / (\overrightarrow{OU} \times \overrightarrow{OS_1}).
\end{aligned}
\end{equation}

\begingroup
\fontsize{6}{6}\selectfont
\setlength{\tabcolsep}{1mm}
\renewcommand{\arraystretch}{1.0} 
\begin{table*}[!t]
\vspace{-0.4cm}
\caption{Performance comparisons on MMLU, ARC-c, TriviaQA and NQ. The best performance is highlighted in \textbf{boldface}, while the second-best performance is \underline{underlined}.}
\vspace{-0.2cm}
\centering
\resizebox{1.0\linewidth}{!}{
\begin{tabular}{ccc|ccc|ccc|ccc|ccc}
\hline
\multirow{3}{*}{\textbf{LLMs}} & \multicolumn{2}{c|}{\textbf{QA Type}} & \multicolumn{6}{c|}{\textbf{MCQA}} & \multicolumn{6}{c}{\textbf{OEQA}} \\
\cline{2-15}
& \multicolumn{2}{c|}{\textbf{Dataset}} & \multicolumn{3}{c|}{\textbf{MMLU (ID)}} & \multicolumn{3}{c|}{\textbf{ARC-c (OOD)}} & \multicolumn{3}{c|}{\textbf{TriviaQA (ID)}} & \multicolumn{3}{c}{\textbf{NQ (OOD)}} \\
\cline{2-15} 
& \multicolumn{2}{c|}{\textbf{Metric}} & $P_c$ & $P_w\downarrow$ & THS$\uparrow$ & $P_c$ & $P_w\downarrow$ & THS$\uparrow$ & $P_c$ & $P_w\downarrow$ & THS$\uparrow$ & $P_c$ & $P_w\downarrow$ & THS$\uparrow$ \\
\hline
\multirow{9}{*}{\shortstack{\textbf{Llama2-7B} \\ \textbf{Chat}}}
& \multirow{5}{*}{\textbf{Baselines}}  & Init-Basic &  45.6 & 52.8 & 00.0 & 53.9 & 46.0 & 00.0 & 54.0 & 46.0 & 00.0 & 28.9 & 71.1 & 00.0 \\ 
&& Init-Refuse & 36.4 & 38.9 & 03.9 & 44.4 & 35.7 & 02.6 & 37.1 & 21.7 & 11.5 & 19.8 & \textbf{34.8} & \textbf{05.6} \\ 
&& Van-Tuning & 46.9 & 53.0 & 01.2 & 54.5 & 45.5 & 01.2 & 55.5 & 44.5 & 03.2 & 23.2 & 76.8 & -0.80 \\ 
&& R-Tuning & 44.5 & 39.6 & 11.3 & 55.8 & 38.1 & 11.1 & 52.2 & 35.9 & 10.0 & 22.6 & 60.9 & -0.22 \\ 
&& CRaFT & 43.9 & 36.4 & 12.5 & 54.7 & 35.9 & 12.6 & 47.8 & 28.1 & 14.8 & 26.7 & 62.0 & 01.5 \\  \cdashline{2-15}
&\textbf{Ours}& \M & 43.5 & \underline{27.1} & \textbf{20.1} & 55.2 & \textbf{26.5} & \textbf{24.2} & 43.6 & \underline{18.4} & \textbf{22.0} & 20.8 & 49.7 & 00.0 \\  \cdashline{2-15}
&\multirow{2}{*}{\textbf{Ablations}}& \texttt{w/o} $\mathbf{O}_{1}$ & 44.7 & 39.8 & 10.3 & 55.4 & 37.9 & 11.0 & 52.4 & 36.5 & 09.6 & 23.9 & 63.5 & -01.9 \\  
&& \texttt{w/o} $\mathbf{O}_{2}$ & 42.8 & \textbf{26.5} & \underline{20.0} & 54.1 & \underline{26.7} & \underline{22.8} & 41.9 & \textbf{18.1} & \underline{20.6} & 20.1 & \underline{48.3} & \underline{00.5} \\  
\hline
\multirow{9}{*}{\shortstack{\textbf{Llama3-8B} \\ \textbf{Instruct}}}
& \multirow{5}{*}{\textbf{Baselines}} & Init-Basic & 66.8 & 33.1 & 00.0 & 80.6 & 19.5 & 00.0 & 66.8 & 33.2 & 00.0 & 40.3 & 59.7 & \underline{00.0} \\ 
&& Init-Refuse & 50.0 & 17.0 & 15.7 & 65.3 & 14.4 & 05.6 & 53.9 & 20.8 & 12.0 & 31.1 & \textbf{38.6} & \textbf{05.0}
\\ 
&& Van-Tuning & 69.5 & 30.5 & 08.0 & 80.3 & 19.7 & -01.3 & 60.0 & 40.0 & -19.0 & 21.0 & 48.5 & -11.7 \\ 
&& R-Tuning & 63.9 & 21.6 & 20.4 & 79.4 & 16.2 & 12.2 & 56.6 & 28.3 & -00.5 & 25.1 & 74.9 & -25.6 \\ 
&& CRaFT & 53.3 & 09.6 & 34.0 & 74.1 & 12.7 & 21.4 & 57.8 & 27.7 & 02.0 & 27.0 & 57.6 & -12.0 \\  \cdashline{2-15}
&\textbf{Ours}& \M & 50.4 & \textbf{06.9} & \textbf{36.4} & 70.2 & \underline{08.7} & \textbf{34.3} & 55.3 & \textbf{18.3} & \textbf{18.5} & 21.9 & \underline{38.8} & -04.4 \\  \cdashline{2-15}
&\multirow{2}{*}{\textbf{Ablations}}& \texttt{w/o} $\mathbf{O}_{1}$ & 64.1 & 21.4 & 20.9 & 79.3 & 16.4 & 11.5 & 57.5 & 28.7 & -00.2 & 25.6 & 75.0 & -25.0 \\  
&& \texttt{w/o} $\mathbf{O}_{2}$ & 49.6 & \underline{07.0} & \underline{35.5} & 69.1 & \textbf{08.6} & \underline{33.6} & 54.3 & \textbf{18.3} & \underline{17.4} & 21.6 & 39.1 & -04.8 \\  
\hline
\end{tabular}}
\label{table:main table}
\end{table*}
\endgroup

\paragraph{Implementation Details.}

In our studies, we utilized LLaMA2-7B-Chat and LLaMA3-8B-Instruct as the initial LLMs \( \theta_0 \). For the MCQA task, we selected 5,000 samples from the MMLU dataset for training purposes, and for the OEQA task, 10,000 samples from TriviaQA were used. With the exception of the Van-Tuning setting, where all samples were kept unchanged, other RAIT settings used a 1:4 ratio of \texttt{ik} samples to \texttt{idk} samples. In the MCQA and OEQA tasks, correctness is obtained using 5-shot and 3-shot setups\footnote{The reasons for using the few-shot settings are outlined in the appendix \ref{A5}}, respectively. More implementation details are listed in Appendix~\ref{app:imple}. In contrast to \cite{zhu2024utilizeflowsteppingriver}, to ensure the fairness of the experiments, we employ LoRA for training across both MCQA and OEQA tasks.

During both training and testing phases, XTuner~\footnote{https://github.com/InternLM/xtuner} was employed for RAIT experiments, which were conducted over 3 epochs with a maximum context length set to 2048. The LoRA~\cite{hulora} was implemented with the parameters: \(r=64\), \(\alpha=16\), dropout rate of 0.1, and a learning rate of \(2 \times 10^{-4}\). For evaluations, the 0-shot approach with greedy decoding was adopted. OpenCompass~\footnote{https://github.com/open-compass/opencompass} is used for all evaluations and correctness calculations. In the \M method, we assigned the hyperparameter \(\mathcal{T}_{\mathcal{C}}\) a value of 0.5 and \(\tau\) a value of 0.05. All experiments were executed on eight NVIDIA A100-80GB GPUs.

\subsection{Experiment Results}
We present the main experimental results, along with an ablation study of \M across various models in Table~\ref{table:main table}. A summary of the key findings is provided below.

\subsubsection{Main Results}
We assess the effectiveness of \M by addressing the challenges \textbf{\textit{C1}} and \textbf{\textit{C2}}, with the corresponding experimental results presented in Table~\ref{table:main table}. \\
\textbf{Comparison based on \textit{C1}:} \textbf{\textit{C1}} relates to the metric \(P_w\), where a lower \(P_w\) indicates better avoidance of hallucinations by the model. As shown in the results, our proposed \M achieves a significantly lower \(P_w\) compared to other baselines, demonstrating its effectiveness in reducing hallucination rates. \\
\textbf{Comparison based on \textit{C2}:} \textbf{\textit{C2}} focuses on minimizing hallucinations while maintaining accuracy, addressing the challenge of over-refusal. 
\M surpasses existing methods in THS score with an average of 3.66.
Specifically, the THS results clearly show that our method significantly outperforms other baselines on both in-domain (ID) and out-of-domain (OOD) settings. For instance, on MMLU dataset, the LLaMA2-7B-Chat model achieves a THS score of 19.3, whereas the best-performing baseline, CRaFT, only reaches 12.5. Moreover, our approach consistently demonstrates superior performance on OOD datasets as well.

\subsubsection{Ablation Study}
We conduct ablation studies to evaluate the contribution of each component in \M, as presented in Table~\ref{table:main table}, using two variants: (1) \M without Refusal Influence, which follows the R-Tuning approach during the dataset distillation phase (denoted as \texttt{w/o} $\mathbf{O}_1$), and (2) \M without Stable Influence, where no weight adjustment is applied to emphasize the importance of \texttt{idk} samples (denoted as \texttt{w/o} $\mathbf{O}_2$). The results indicate that each component contributes positively to the overall performance of \M and the removal of any component leads to a noticeable decline in effectiveness. Specifically, replacing Refusal Influence-based dataset distillation with other baselines results in a significant increase in hallucination rate, underscoring the importance of Refusal Influence in addressing \textbf{\textit{C1}}. Additionally, the use of Stable Influence helps reduce over-refusal while maintaining a stable hallucination rate, effectively addressing the challenges posed by \textbf{\textit{C2}}. In addition, we conducted sensitivity experiments, the details of which can be found in the appendix \ref{A6}.


\begin{figure}[t]
    \centering
    \includegraphics[width=1.0\linewidth]{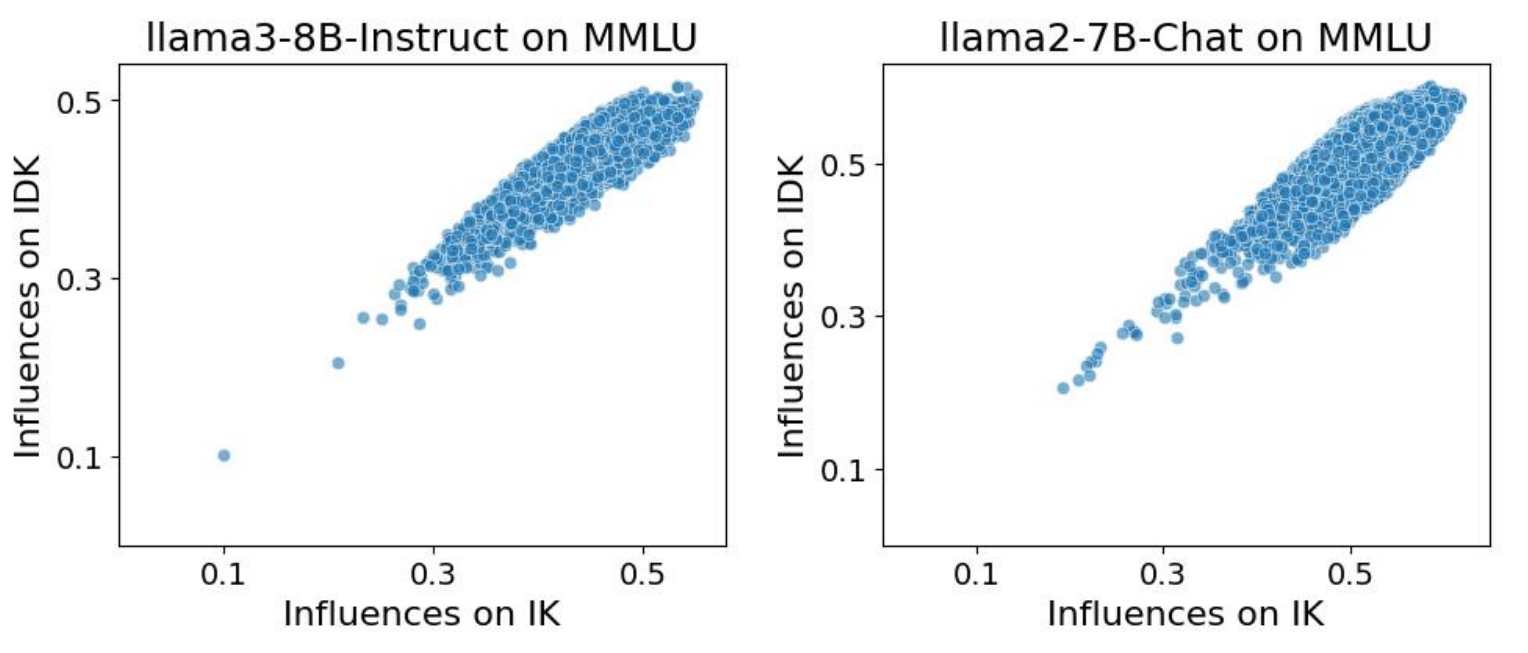}
    \caption{Relationship between \(\mathcal{I}^{\text{ref}}\) and \(\mathcal{I}^{\text{over}}\) in MMLU performance on LLaMA2-7B-Chat and LLaMA3-8B-Instruct.}
    \label{fig:analysis}
\end{figure}

\subsection{Analysis}
\textbf{The selection of \texttt{ik} samples is crucial.} Our analysis and experiments primarily focus on optimizing the selection of \texttt{idk} samples. However, the selection of \texttt{ik} samples is also crucial. We employed three different strategies: \texttt{ik-random}, where data is randomly selected from $D_{\text{ik}}$; \texttt{ik-bottom}, where the data with the lowest correctness from $D_{\text{ik}}$ is selected; and \texttt{ik-top}, the method used in \M, where the data with the highest correctness from $D_{\text{ik}}$ is chosen. We used the MMLU (ID) and ARC-c (OOD) datasets and conducted experiments with the LLaMA3-8B-Instruct model. The results are shown in Table 3. When using either the \texttt{ik-bottom} or \texttt{ik-random} methods, the model's hallucination reduction does not improve, and the refusal rate remains low. We believe the potential reason for this is that the \texttt{ik} samples selected by these methods may share similar characteristics with the \texttt{idk} samples, but different supervision signals were applied during the SFT process. This weakens the model’s ability to learn effective refusals. In contrast, the \texttt{ik-top} strategy used in \M helps to distinctly separate the features of the two types of samples, addressing the static conflict mentioned in \cite{zhu2024utilizeflowsteppingriver}.

\begingroup
\fontsize{6}{6}\selectfont
\setlength{\tabcolsep}{1mm}
\renewcommand{\arraystretch}{1.2} 
\label{table:Analysis}
\begin{table}[!t]
\small  
\vspace{-0.4cm}
\caption{Performance comparisons on MMLU and ARC-c for different \texttt{ik} selection methods on LLaMA3-8B-Instruct.}
\vspace{-0.2cm}
\centering
\begin{tabular}{c|ccc|ccc}
\hline
\textbf{Dataset} & \multicolumn{3}{c|}{\textbf{MMLU (ID)}} & \multicolumn{3}{c}{\textbf{ARC-c (OOD)}} \\
\cline{1-7} 
\textbf{Metric} & $P_c$ & $P_w\downarrow$ & THS$\uparrow$ & $P_c$ & $P_w\downarrow$ & THS$\uparrow$ \\
\hline
\texttt{ik-top} & 50.4 & 06.9 & 36.4 & 70.2 & 08.7 & 34.3 \\ 
\texttt{ik-random} & 61.4 & 20.5 & 20.0 & 78.7 & 15.6 & 14.2 \\ 
\texttt{ik-bottom} & 64.0 & 25.3 & 12.9 & 79.7 & 19.0 & -00.2 \\ 
\hline
\end{tabular}
\end{table}

\endgroup

\textbf{Over-Refusal can only be alleviated, but not completely eliminated.}
During the RAIT process, we observed and analyzed the \texttt{idk} influence (corresponding to $O_{2}$) of \texttt{idk} samples on $D_{\text{ik}}$ and $D_{\text{idk}}$ using the LLaMA2-7B-Chat and LLaMA3-8B-Instruct model on the MMLU dataset. As shown in Figure~\ref{fig:analysis}, we identified a strong correlation between the two, with a Pearson Correlation Coefficient of 0.886. This correlation may be a contributing factor to the occurrence of Over-Refusal. While our proposed method, as indicated in Table~\ref{table:main table}, cannot fully eliminate Over-Refusal due to certain limitations, it significantly mitigates the issue.

\section{Conclusion and Future Work}
\label{sec:Conclusion}
In this paper, we present \M, a Gradient-based Refusal-Aware Instruction Tuning Framework, which addresses the critical challenge of over-refusal in existing RAIT approaches. By leveraging insights derived from a gradient perspective, \M effectively distills refusal-aware datasets and incorporates an adaptive weighting mechanism during fine-tuning. Our experimental results demonstrate that \M not only mitigates hallucinations but also enhances the reliability and accuracy of LLM outputs.
Looking ahead, we aim to further investigate two key avenues of research. First, we plan to explore the dynamic trajectory influence of gradients throughout the RAIT process, which could provide deeper insights into how various training samples impact LLM refusal behavior. Second, we intend to examine the role of \M in detecting knowledge boundaries within LLMs, focusing on its potential contributions to enhancing LLM safety.

\section*{Limitations}
\label{latex/limitation}
While our work has yielded promising results, it is important to recognize several limitations. First, the \M framework currently treats the training process as static, rather than incorporating the dynamic influence of gradient trajectories throughout the RAIT process. Additionally, the \texttt{idk} and \texttt{ik} sets are divided through a straightforward query of the LLMs; future work could explore ways to leverage \M for more nuanced identification of knowledge boundaries within LLMs for splitting. Finally, although \M has demonstrated strong generalizability across various evaluation datasets, expanding the dataset range to include a more diverse set of high-quality resources could enhance the robustness and versatility of the framework.

\section*{Acknowledgments}
This research was supported by Shanghai Artificial Intelligence Laboratory. 

\bibliography{custom}

\appendix

\section{Appendix}
\label{sec:appendix}

\subsection{Theoretical Analysis Details}
\subsubsection{More Proof on $\mathbf{O}_1$}
\label{app:More Proof on O1}

\begin{equation}
\scriptsize
\setlength{\jot}{5pt} 
\label{eq:loss_decomposition}
\begin{aligned}
&\mathbb{E}_{x^{u}_{\text{idk}} \sim D_{\text{idk}}} \left[ \Delta \mathcal{L}(x^{u}_{\text{idk}}, y^{u}_{\text{idk}}; \theta) \right] \\
\equiv & -\eta_t \Bigl \langle  
\mathbb{E}_{x^{u}_{\text{idk}} \sim D_{\text{idk}}} \left[ \nabla \mathcal{L}(x^{u}_{\text{idk}}, y^{u}_{\text{idk}}; \theta) \right], \\
&\quad \mathbb{E}_{x^{o} \sim D} \left[ \nabla \mathcal{L}(x^{o}, y^{o}; \theta) \right] \Bigr \rangle + \mathcal{O}(\eta^2) \\ 
= & -\eta_t \Bigl \langle  
\mathbb{E}_{x^{u}_{\text{idk}} \sim D_{\text{idk}}} \left[ \nabla \mathcal{L}(x^{u}_{\text{idk}}, y^{u}_{\text{idk}}; \theta) \right], \\
&\quad \Bigl( \mathbb{E}_{x^{o}_{\text{ik}} \sim D_{\text{ik}}} \left[ \nabla \mathcal{L}(x^{o}_{\text{ik}}, y^{o}_{\text{ik}}; \theta) \right] + \\
&\quad \mathbb{E}_{x^{o}_{\text{idk}} \sim D_{\text{idk}}} \left[ \nabla \mathcal{L}(x^{o}_{\text{idk}}, y^{o}_{\text{idk}}; \theta) \right] 
\Bigr) \Bigr \rangle + \mathcal{O}(\eta^2) \\
\approx & -\eta_t \mathbb{E}_{x^{u}_{\text{idk}} \sim D_{\text{idk}}} \left[ \nabla \mathcal{L}(x^{u}_{\text{idk}}, y^{u}_{\text{idk}}; \theta) \right] \cdot \\
&\quad \mathbb{E}_{x^{o}_{\text{idk}} \sim D_{\text{idk}}} \left[ \nabla \mathcal{L}(x^{o}_{\text{idk}}, y^{o}_{\text{idk}}; \theta) \right] \\
= & -\eta_t 
\mathbb{E}_{(x^{u}_{\text{idk}},x^o_{\text{idk}}) \sim D_{\text{idk}}} \left[ \nabla \mathcal{L}(x^{u}_{\text{idk}}, y^{u}_{\text{idk}}; \theta) \cdot \right. \\
&\quad \left. \nabla \mathcal{L} (x^{o}_{\text{idk}},y^{o}_{\text{idk}}; \theta) \right]. \\
& = - \mathbb{E}_{(x^{u}_{idk}, x^o_{idk}) \sim D_{idk}} \left[ \mathcal{I}(x^{o}_{idk}, y^{o}_{idk}, x^{u}_{idk}, y^{u}_{idk}; \theta) \right]
\end{aligned}
\end{equation}

\subsubsection{More Proof on $\mathbf{O}_2$}
\label{app:More Proof on O2}
RAIT can lead to the phenomenon of over-refusal, where the model refuses to answer questions it is inherently capable of addressing, thereby resulting in a decrease in accuracy. Accordingly, for an unlabeled input query $x_{ik}^u$, the output responses will shift from $y_{ik}$ to $y_{idk}$. Assuming the use of a symmetric loss function, the difference in the loss function values for the same input $x_{ik}$ with target labels $y_{ik}$ and $y_{idk}$ is approximately opposite in sign: $\Delta L(x_{ik}, y_{ik}) \approx -\Delta L(x_{ik}, y_{idk})$. Therefore, the proof is as follows:
\begin{equation}
\scriptsize
\setlength{\jot}{5pt} 
\label{eq:loss_decomposition}
\begin{aligned}
&\mathbb{E}_{x^{u}_{\text{idk}} \sim D_{\text{idk}}} \left[ \Delta \mathcal{L}(x^{u}_{\text{idk}}, y^{u}_{\text{idk}}; \theta) \right] 
+ \mathbb{E}_{x^{u}_{\text{ik}} \sim D_{\text{ik}}} \left[ \Delta \mathcal{L}(x^{u}_{\text{ik}}, y^{u}_{\text{ik}}; \theta) \right] \\ 
\approx & ~\mathbb{E}_{x^{u}_{\text{idk}} \sim D_{\text{idk}}} \left[ \Delta \mathcal{L}(x^{u}_{\text{idk}}, y^{u}_{\text{idk}}; \theta) \right] 
- \mathbb{E}_{x^{u}_{\text{ik}} \sim D_{\text{ik}}} \left[ \Delta \mathcal{L}(x^{u}_{\text{ik}}, y^{u}_{\text{idk}}; \theta) \right] \\ 
\equiv & -\eta_t \Bigl \langle  
\mathbb{E}_{x^{u}_{\text{idk}} \sim D_{\text{idk}}} \left[ \nabla \mathcal{L}(x^{u}_{\text{idk}}, y^{u}_{\text{idk}}; \theta) \right], \\
&\quad \mathbb{E}_{x^{o} \sim D} \left[ \nabla \mathcal{L}(x^{o}, y^{o}; \theta) \right] \Bigr \rangle + \eta_t \Bigl \langle  
\mathbb{E}_{x^{u}_{\text{ik}} \sim D_{\text{ik}}} \left[ \nabla \mathcal{L}(x^{u}_{\text{ik}}, y^{u}_{\text{idk}}; \theta) \right], \\
&\quad \mathbb{E}_{x^{o} \sim D} \left[ \nabla \mathcal{L}(x^{o}, y^{o}; \theta) \right] \Bigr \rangle + \mathcal{O}(\eta^2) \\ 
= & - \eta_t \Bigg\langle 
\Bigg\{ \mathbb{E}_{x^{u}_{\text{idk}} \sim D_{\text{idk}}} 
\left[ \nabla \mathcal{L}(x^{u}_{\text{idk}}, y^{u}_{\text{idk}}; \theta) \right] \\
& \quad - \mathbb{E}_{x^{u}_{\text{ik}} \sim D_{\text{ik}}} 
\left[ \nabla \mathcal{L}(x^{u}_{\text{ik}}, y^{u}_{\text{ik}}; \theta) \right] 
\Bigg\}, \\ 
& \quad 
\Bigg\{ \mathbb{E}_{x^{o}_{\text{idk}} \sim D_{\text{idk}}} 
\left[ \nabla \mathcal{L}(x^{o}_{\text{idk}}, y^{o}_{\text{idk}}; \theta) \right] 
+ \mathbb{E}_{x^{o}_{\text{ik}} \sim D_{\text{ik}}} 
\left[ \nabla \mathcal{L}(x^{o}_{\text{ik}}, y^{o}_{\text{ik}}; \theta) \right] 
\Bigg\} 
\Bigg\rangle \\ 
& \quad + \mathcal{O}(\eta^2)
 \\
\approx & - \eta_t \left \langle  
\left \{
\mathbb{E}_{x^{u}_{\text{idk}} \sim D_{\text{idk}}} \left[ \nabla \mathcal{L}(x^{u}_{\text{idk}}, y^{u}_{\text{idk}}; \theta) \right] 
\right. \right. \\
& \quad \left. - \mathbb{E}_{x^{u}_{\text{ik}} \sim D_{\text{ik}}} \left[ \nabla \mathcal{L}(x^{u}_{\text{ik}}, y^{u}_{\text{idk}}; \theta) \right] \right \},  \mathbb{E}_{x^{o}_{\text{idk}} \sim D_{\text{idk}}} \left[ \nabla \mathcal{L}(x^{o}_{\text{idk}}, y^{o}_{\text{idk}}; \theta) \right] \Big \rangle \\
= & -\eta_t \left \{
\mathbb{E}_{(x^{u}_{\text{idk}}, x^o_{\text{idk}}) \sim D_{\text{idk}}} \left[ \nabla \mathcal{L}(x^{u}_{\text{idk}}, y^{u}_{\text{idk}}; \theta) \cdot \nabla \mathcal{L}(x^{o}_{\text{idk}}, y^{o}_{\text{idk}}; \theta) \right] \right. \\
&\quad - \left.
\mathbb{E}_{x^{u}_{\text{ik}} \sim D_{\text{ik}}, x^{o}_{\text{idk}} \sim D_{\text{idk}}} \left[ \nabla \mathcal{L}(x^{u}_{\text{ik}}, y^{u}_{\text{idk}}; \theta) \cdot \nabla \mathcal{L}(x^{o}_{\text{idk}}, y^{o}_{\text{idk}}; \theta) \right] \right \} \\
= & - \left \{ \mathbb{E} _{x^{u} _{\text{idk}}, x^o _{\text{idk}} \sim D _{\text{idk}}} \left[ \mathcal{I}(x^{o} _{\text{idk}}, y^{o} _{\text{idk}}, x^{u} _{\text{idk}}, y^{u} _{\text{idk}}; \theta) \right] \right. \\
& \quad - \left . \mathbb{E} _{x^{u} _{\text{ik}} \sim D _{\text{ik}}, x^{o} _{\text{idk}} \sim D _{\text{idk}}} \left[ \mathcal{I}(x^{o} _{\text{idk}}, y^{o} _{\text{idk}}, x^{u} _{\text{ik}}, y^{u} _{\text{idk}}; \theta) \right] \right \}
\end{aligned}
\end{equation}

Here, the first approximation is transformed as follows:
\begin{itemize}[leftmargin=*]
    \item RAIT can lead to the phenomenon of over-refusal, where the model refuses to answer questions it is inherently capable of addressing, thereby resulting in a decrease in accuracy. Accordingly, for an unlabeled input query $x_{ik}^u$, the output responses will shift from $y_{ik}$ to $y_{idk}$.
    \item Assuming the use of a symmetric loss function, the difference in the loss function values for the same input $x_{ik}$ with target labels $y_{ik}$ and $y_{idk}$ is approximately opposite in sign: 
    $$
    \Delta \mathcal{L}(x_{ik}, y_{ik}) \approx -\Delta \mathcal{L}(x_{ik}, y_{idk}).
    $$
\end{itemize}

\subsection{Reasons for Choosing Influence Formula}
\label{app:Reasons for Choosing Influence Formula}
The reasons for choosing the influence formula of Equations (4) and (5) instead of the optimizing of Equation (2):
\begin{itemize}[leftmargin=*]
    \item The left-hand term, $\Delta \mathcal{L}$, in Equations (4) and (5) represents our optimization objective on the test set. Our goal is to identify the training samples that minimize this $\Delta \mathcal{L}$.
    \item \textbf{Computational complexity without approximation is $O(m \cdot n)$}: Directly computing the contribution of each training sample to $\Delta \mathcal{L}$ is computationally expensive. Assuming the training set size is $m$ and the test set size is $n$, we need to compute the loss change for \textbf{each test sample} after training on \textbf{each individual training sample}, with time complexity of $O(m \cdot n)$.
    \item \textbf{Computational complexity is reduced to $O(m + n)$ after approximation}: Inspired by \cite{Pruthi_Liu_Kale_Sundararajan_2020, Xia_Malladi_Gururangan_Arora_Chen}, we approximate $\Delta \mathcal{L}$ using the influence function. Using the influence function to approximate $\Delta \mathcal{L}$ only requires computing the gradients of each training sample and test sample without repeated calculations. Its time complexity is $O(m + n)$, which significantly reduces computational overhead. This approximation arises from omitting the higher-order terms in the Taylor expansion during the derivation of the influence function. 
\end{itemize}

\subsection{Orthogonal Experiment}
\label{subsec:orthogonal_experiment}
\begin{figure*}[t]
  \centering
  \includegraphics[width=1.0\linewidth]{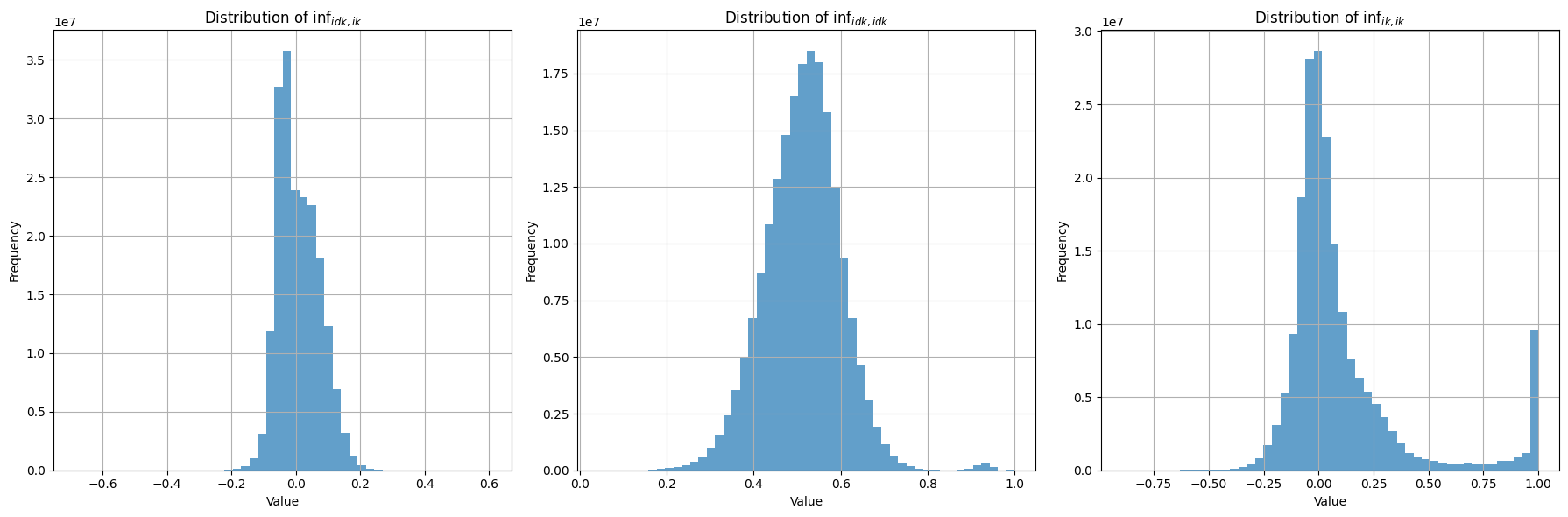}
  \caption{The overview of our proposed \M.
  }
  \label{fig:orth}
\end{figure*}

In our experiments, we observed that the gradient distributions for \texttt{idk} and \texttt{ik} are nearly orthogonal, as illustrated in Figure~\ref{fig:orth}. Through experiments conducted on the Llama2-7B-Chat model and the MMLU dataset, we found that the inner product distribution between \texttt{idk} gradients and \texttt{ik} gradients is centered around zero, with the computed value being
$
\Bigl \langle \mathbb{E}_{x } \left[ \nabla \mathcal{L}(x,y_{\text{idk}}; \theta) \right], 
\mathbb{E}_{x} \left[ \nabla \mathcal{L}(x,y_{\text{ik}}; \theta) \right] \Bigr \rangle = 0.008.
$
In contrast, the inner product of \textit{ik} with itself is significantly larger,
$
\Bigl \langle \mathbb{E}_{x } \left[ \nabla \mathcal{L}(x,y_{\text{ik}}; \theta) \right], 
\mathbb{E}_{x} \left[ \nabla \mathcal{L}(x,y_{\text{ik}}; \theta) \right] \Bigr \rangle = 0.103,
$
and the inner product of \texttt{idk} with itself is even greater,
$
\Bigl \langle \mathbb{E}_{x } \left[ \nabla \mathcal{L}(x,y_{\text{idk}}; \theta) \right], 
\mathbb{E}_{x} \left[ \nabla \mathcal{L}(x,y_{\text{idk}}; \theta) \right] \Bigr \rangle = 0.513.
$

Our explanation for this phenomenon includes the following points:
\begin{itemize}[leftmargin=*]
    \item \texttt{ik} samples and \texttt{idk} samples train different capabilities of the LLM: From the perspective of the internal knowledge of the LLM, \texttt{ik} samples help the LLM ``\textbf{learn knowledge}'', while \texttt{idk} samples grant the LLM the ability to ``\textbf{reflect on self-knowledge}'', that is, to predict the boundaries of its own knowledge.
    \item \textbf{The different capabilities of the LLM are usually associated with different regions or activation patterns of the transformer}: For example, \citep{dai2022knowledgeneuronspretrainedtransformers,yu2024neuronlevelknowledgeattributionlarge} indicates that knowledge is primarily stored in specific neurons of the LLM, while some studies show that different attention heads perform different functions, such as the Successor Head~\cite{gould2023successorheadsrecurringinterpretable} and the Induction Head~\cite{ren2024identifyingsemanticinductionheads}. Additionally, the multilingual capabilities of the LLM mainly depend on the layers near the input and output ends of the transformer, rather than the middle layers~\cite{wendler2024llamasworkenglishlatent}.
    \item The abilities to ``learn knowledge'' (\texttt{ik} samples) and ``reflect on self-knowledge'' (\texttt{idk} samples) correspond to different regions or activation patterns of the LLM transformer. Therefore, during the training process, the gradients of \texttt{ik} samples and \texttt{idk} samples act on different regions or activation patterns of the transformer respectively. This helps explain why the gradient distributions of \texttt{ik} samples and \texttt{idk} samples exhibit a state of near orthogonality.
\end{itemize}

\subsection{Prompts in \M}
\label{app:imple}
\subsubsection{Prompts for Getting Correctness.}
Prompts for getting correctness on MMLU and TriviaQA datasets are shown in Table 4 and Table 5. They use 5-shot and 3-shot settings respectively.

\begin{table}[]
\centering

\begin{tcolorbox}[title={In-Context Examples}, colback=white, coltitle=black, colbacktitle=white!0]
There is a single choice question about \{\textit{Task}\}. Answer the question by replying A, B, C or D.\\
\textbf{Question}: \{\textit{Question1}\} \\
\textbf{A}. \{\textit{Content\_of\_A1}\}\\
\textbf{B}. \{\textit{Content\_of\_B1}\}\\
\textbf{C}. \{\textit{Content\_of\_C1}\}\\
\textbf{D}. \{\textit{Content\_of\_D1}\}\\
\textbf{Answer}: \{\textit{Answer1}\}\\
\\
There is a single choice question about \{\textit{Task}\}. Answer the question by replying A, B, C or D.\\
\textbf{Question}: \{\textit{Question2}\} \\
\textbf{A}. \{\textit{Content\_of\_A2}\}\\
\textbf{B}. \{\textit{Content\_of\_B2}\}\\
\textbf{C}. \{\textit{Content\_of\_C2}\}\\
\textbf{D}. \{\textit{Content\_of\_D2}\}\\
\textbf{Answer}: \{\textit{Answer2}\}\\
\\
$\dots$
\\
There is a single choice question about \{\textit{Task}\}. Answer the question by replying A, B, C or D.\\
\textbf{Question}: \{\textit{Question5}\} \\
\textbf{A}. \{\textit{Content\_of\_A5}\}\\
\textbf{B}. \{\textit{Content\_of\_B5}\}\\
\textbf{C}. \{\textit{Content\_of\_C5}\}\\
\textbf{D}. \{\textit{Content\_of\_D5}\}\\
\textbf{Answer}: \{\textit{Answer5}\}\\
\end{tcolorbox}

\begin{tcolorbox}[title={Instruction}, colback=white, coltitle=black, colbacktitle=white!0]
There is a single choice question about \{\textit{Task}\}. Answer the question by replying A, B, C or D.\\
\textbf{Question}: \{\textit{Question}\} \\
\textbf{A}. \{\textit{Content\_of\_A}\}\\
\textbf{B}. \{\textit{Content\_of\_B}\}\\
\textbf{C}. \{\textit{Content\_of\_C}\}\\
\textbf{D}. \{\textit{Content\_of\_D}\}\\
\textbf{Answer}:
\end{tcolorbox}

\caption{The Prompt Template for Knowledge State Query on MMLU. The Italic \{\textit{text}\} in Curly Braces Represents Variables That Need To be Replaced.}
\label{table:prompt_kq_MMLU}

\end{table}

\begin{table}[]
    \centering
    
    \begin{tcolorbox}[title={In-Context Examples}, colback=white, coltitle=black, colbacktitle=white!0]
    Answer the following question as simple as possible.\\
    \textbf{Question}: \{\textit{Question1}\}\\
    \textbf{Answer}: \{\textit{Answer1}\}\\
    \\
    Answer the following question as simple as possible.\\
    \textbf{Question}: \{\textit{Question2}\}\\
    \textbf{Answer}: \{\textit{Answer2}\}\\
    \\
    Answer the following question as simple as possible.\\
    \textbf{Question}: \{\textit{Question3}\}\\
    \textbf{Answer}: \{\textit{Answer3}\}\\
    \end{tcolorbox}
    
    \begin{tcolorbox}[title={Instruction}, colback=white, coltitle=black, colbacktitle=white!0]
    Answer the following question as simple as possible.\\
    \textbf{Question}: \{\textit{Question}\}\\
    \textbf{Answer}: 
    \end{tcolorbox}
    
    \caption{The Prompt Template for Knowledge State Query on TriviaQA. The Italic \{\textit{text}\} in Curly Braces Represents Variables That Need To be Replaced.}
    \label{table:prompt_kq_triviaqa}
    
    \end{table}

\subsubsection{Prompts for training.}
For the \textit{Van-Tuning}, we use the \textit{basic} prompt as shown in Table 4 and Table 5 without in-context example. All other experiments use the \textit{refuse} prompt as shown in Table 6 and Table 7. Loss is only computed on the target answer $\{\text{answer}_\text{rait}\}$.

\begin{table}[]
\centering

\begin{tcolorbox}[title={Instruction}, colback=white, coltitle=black, colbacktitle=white!0]
There is a single choice question about \{\textit{Task}\}. If you know the answer, please directly respond with the correct answer A, B, C, or D. If you do not know the answer, please respond with ``I don't know.".\\
\textbf{Question}:\{\textit{Question}\} \\
\textbf{A}. \{\textit{Content\_of\_A}\}\\
\textbf{B}. \{\textit{Content\_of\_B}\}\\
\textbf{C}. \{\textit{Content\_of\_C}\}\\
\textbf{D}. \{\textit{Content\_of\_D}\}\\
\textbf{Answer}: \{$Answer_{\text{rait}}$\} 
\end{tcolorbox}

\caption{The \textbf{REFUSE} Prompt Template for \textbf{Training} on MMLU. The The Italic \{\textit{text}\} in Curly Braces Represents Variables That Need To be Replaced.}
\label{table:prompt_training_refuse_mmlu}

\end{table}

\begin{table}[]
\centering

\begin{tcolorbox}[title={Instruction}, colback=white, coltitle=black, colbacktitle=white!0]
Answer the following question, and if you don't know the answer, only reply with ``I don't know'':\{\textit{Question}\}\\
\{$Answer_{\text{rait}}$\} 
\end{tcolorbox}

\caption{The \textbf{REFUSE} Prompt Template for \textbf{Training} on TriviaQA. The Italic \{\textit{text}\} in Curly Braces Represents Variables That Need To be Replaced.}
\label{table:prompt_training_refuse_triviaqa}

\end{table}

\subsubsection{Prompts for evaluation.}
The \textit{Init-Basic} method uses the original question format for evaluation, without any prior instructions. For the other methods, the evaluation prompts are shown in Tables 8 and 9.

\begin{table}[]
\centering

\begin{tcolorbox}[title={Instruction}, colback=white, coltitle=black, colbacktitle=white!0]
There is a single choice question about \{\textit{Task}\}. If you know the answer, please directly respond with the correct answer A, B, C, or D. If you do not know the answer, please respond with ``I don't know.''.\\
\textbf{Question}:\{\textit{Question}\} \\
\textbf{A}. \{\textit{Content\_of\_A}\}\\
\textbf{B}. \{\textit{Content\_of\_B}\}\\
\textbf{C}. \{\textit{Content\_of\_C}\}\\
\textbf{D}. \{\textit{Content\_of\_D}\}\\
\textbf{Answer}: 
\end{tcolorbox}

\caption{The \textbf{REFUSE} Prompt Template for \textbf{Evaluation} on MMLU. The Italic \{\textit{text}\} in Curly Braces Represents Variables That Need To be Replaced.}
\label{table:prompt_eval_refuse_mmlu}

\end{table}

\begin{table}[]
\centering

\begin{tcolorbox}[title={Instruction}, colback=white, coltitle=black, colbacktitle=white!0]
Answer the following question, and if you don't know the answer, only reply with ``I don't know": \{\textit{Question}\}
\end{tcolorbox}

\caption{The \textbf{REFUSE} Prompt Template for \textbf{Evaluation} on TriviaQA. The Italic \{\textit{text}\} in Curly Braces Represents Variables That Need To be Replaced.}
\label{table:prompt_eval_refuse_triviaqa}

\end{table}

\subsection{Reasons for Choosing Few-Shot Setting}
\label{A5}
We use few-shot prompting to calculate the correctness of LLMs, aiming to ensure that \textbf{LLMs strictly follow instructions to achieve the following objectives}:
\begin{itemize}[leftmargin=*]
    \item \textbf{Easy-to-parse response}: We extract answers from the response using rules and then compare them with the gt to calculate correctness. Both research~\cite{xfinder} and our practice show that accurately extracting answers is challenging. Therefore, we provide examples of easy-to-parse answers through few-shot examples to ensure compliance with instructions.
    \item \textbf{Clear answers}: Existing chat LLMs often have a certain level of self-awareness and may choose not to answer or give vague responses like ``I'm not sure'' or ``for reference only'' when facing uncertain questions. However, we expect the LLM to answer the query directly, regardless of whether the answer is correct.
    \item \textbf{Concise answers to reduce inference costs}: In a 0-shot scenario, LLMs tend to produce longer responses (chain of thought processes or detailed descriptions). Using few-shot prompts helps obtain concise answers.
\end{itemize}
\textbf{Although descriptive instructions can be added to 0-shot prompts to require LLMs to meet the above standards, LLMs generally do not strictly follow them.} Furthermore, we believe that \textbf{the risk of introducing knowledge by few-shot prompting is minimal}. When choosing few-shot samples, we select from \textbf{different dataset splits}: for MCQA, test samples are from the MMLU test split and few-shot samples from the val split; for OEQA, test samples are from the triviaqa train split, with few-shot samples from the dev split.

\subsection{Sensitive Experiment}
\label{A6}
\textbf{Effect of temprature on \texttt{idk} samples' weight.}
We analyzed the Stable Influence of the samples discussed in Stage 3 of \M and found that their values were relatively small. As a result, it was necessary to adjust the temperature to better normalize the weight assigned to each sample. To explore the specific effects of these adjustments, we conducted experiments using the MMLU dataset and the LLaMA3-8B-Instruct model. As shown in Table~\ref{tab:tem_results}, the weight of the \texttt{idk} samples only becomes effective when \( \tau \) is within a reasonable range. When \( \tau \) is set to 1, the weights of all samples remain almost unchanged, being equal to 1.

\begin{table}[ht]
    \centering
    \begin{tabular}{c|cccccc}
        \hline
        \( \tau \) & 0.01 & 0.05 & 0.1 & 0.2 & 0.5 & 1.0 \\ \hline
        THS & 36.6 & 36.4 & 36.5 & 35.9 & 35.5 & 35.5 \\ \hline
    \end{tabular}
    \caption{Effect of temperature $\tau$ on \texttt{idk} samples' weights.}
    \label{tab:tem_results}
\end{table}

\begin{table}[ht]
    \centering
    \begin{tabular}{c|ccccc}
        \hline
        \( \mathcal{T}_{\mathcal{C}} \) & 0.3 & 0.4 & 0.5 & 0.6 & 0.7 \\ \hline
        THS & 37.4 & 36.7 & 36.4 & 36.4 & 36.3 \\ \hline
    \end{tabular}
    \caption{Effect of correctness threshold.}
    \label{tab:ths_tc_results}
\end{table}

\textbf{Effect of correctness threshold.}
In our experiments, we set the correctness threshold \( \mathcal{T}_{\mathcal{C}} \) to 0.5. We conducted detailed experiments to determine this threshold, and as shown in Table~\ref{tab:ths_tc_results}, the model's performance is not highly sensitive to the choice of \( \mathcal{T}_{\mathcal{C}} \). Within a reasonable range, our method consistently delivers strong results. All experiments were performed on the MMLU dataset using the LLaMA3-8B-Instruct model.


\end{document}